\documentclass[letterpaper]{article} % DO NOT CHANGE THIS
\usepackage{aaai25}  % DO NOT CHANGE THIS
\usepackage{times}  % DO NOT CHANGE THIS
\usepackage{helvet}  % DO NOT CHANGE THIS
\usepackage{courier}  % DO NOT CHANGE THIS
\usepackage[hyphens]{url}  % DO NOT CHANGE THIS
\usepackage{graphicx} % DO NOT CHANGE THIS
\usepackage{natbib}  % DO NOT CHANGE THIS AND DO NOT ADD ANY OPTIONS TO IT
\usepackage{caption} % DO NOT CHANGE THIS AND DO NOT ADD ANY OPTIONS TO IT
\usepackage{amsmath}
\usepackage{bbding}
\usepackage{booktabs}
\usepackage{multirow}
\usepackage{makecell}
\usepackage[utf8]{inputenc}   %%%%%%%%%%%%%%%%%%%%%%%%%%%%%%%%%%%%%%%%%%%%%%
\usepackage{algorithm}
\usepackage{algorithmic}

\usepackage{newfloat}
\usepackage{listings}
\DeclareCaptionStyle{ruled}{labelfont=normalfont,labelsep=colon,strut=off} % DO NOT CHANGE THIS
\floatstyle{ruled}
\newfloat{listing}{tb}{lst}{}
\floatname{listing}{Listing}
\title{Pinwheel-shaped Convolution and Scale-based Dynamic Loss for \\ Infrared Small Target Detection}
\author{
    Jiangnan Yang\textsuperscript{\rm 1},
    Shuangli Liu\textsuperscript{\rm 1}\thanks{Corresponding author: Shuangli Liu.},
    Jingjun Wu\textsuperscript{\rm 2},
    Xinyu Su\textsuperscript{\rm 1},
    Nan Hai\textsuperscript{\rm 1},
    Xueli Huang\textsuperscript{\rm 1}
}
\affiliations{
    \textsuperscript{\rm 1}School of Information and Engineering, Southwest University of Science and Technology\\
    \textsuperscript{\rm 2}School of Electronic and Optical Engineering, Nanjing University of Science and Technology\\
    slliu@swust.edu.cn
}

\usepackage{bibentry}
\begin{document}

\maketitle

\begin{abstract}
	These recent years have witnessed that convolutional neural network (CNN)-based methods for detecting infrared small targets have achieved outstanding performance. However, these methods typically employ standard convolutions, neglecting to consider the spatial characteristics of the pixel distribution of infrared small targets. Therefore, we propose a novel pinwheel-shaped convolution (PConv) as a replacement for standard convolutions in the lower layers of the backbone network. PConv better aligns with the pixel Gaussian spatial distribution of dim small targets, enhances feature extraction, significantly increases the receptive field, and introduces only a minimal increase in parameters.
	Additionally, while recent loss functions combine scale and location losses, they do not adequately account for the varying sensitivity of these losses across different target scales, limiting detection performance on dim-small targets. To overcome this, we propose a scale-based dynamic (SD) Loss that dynamically adjusts the influence of scale and location losses based on target size, improving the network’s ability to detect targets of varying scales.
	We construct a new benchmark, SIRST-UAVB, which is the largest and most challenging dataset to date for real-shot single-frame infrared small target detection. Lastly, by integrating PConv and SD Loss into the latest small target detection algorithms, we achieved significant performance improvements on IRSTD-1K and our SIRST-UAVB dataset, validating the effectiveness and generalizability of our approach.
	\begin{links}
		\link{Code}{https://github.com/JN-Yang/PConv-SDloss-Data}
	\end{links}
\end{abstract}

\section{Introduction}
Infrared small target detection and segmentation (IRSTDS) is widely used in both military and civilian fields due to its advantages, such as thermal sensitivity, nighttime operability, passive radiation, and strong anti-interference capabilities \cite{1,2}. These applications include aircraft and bird warning systems \cite{3,38}, missile guidance systems \cite{4}, and sea rescue operations \cite{5,6}. Typically, these tasks require mid-to-long-range target observation, resulting in small target imaging \cite{2}. As the infrared radiation received by the camera decreases with distance, targets often appear dim with low signal-to-noise ratio (SNR) and signal-to-clutter ratio (SCR), and lack texture information. Additionally, varying distances alter target size and shape. Complex backgrounds, such as buildings, clouds, or vegetation, further obscure targets \cite{7}.

\begin{figure}[t]
	\centering
	\includegraphics[width=\linewidth]{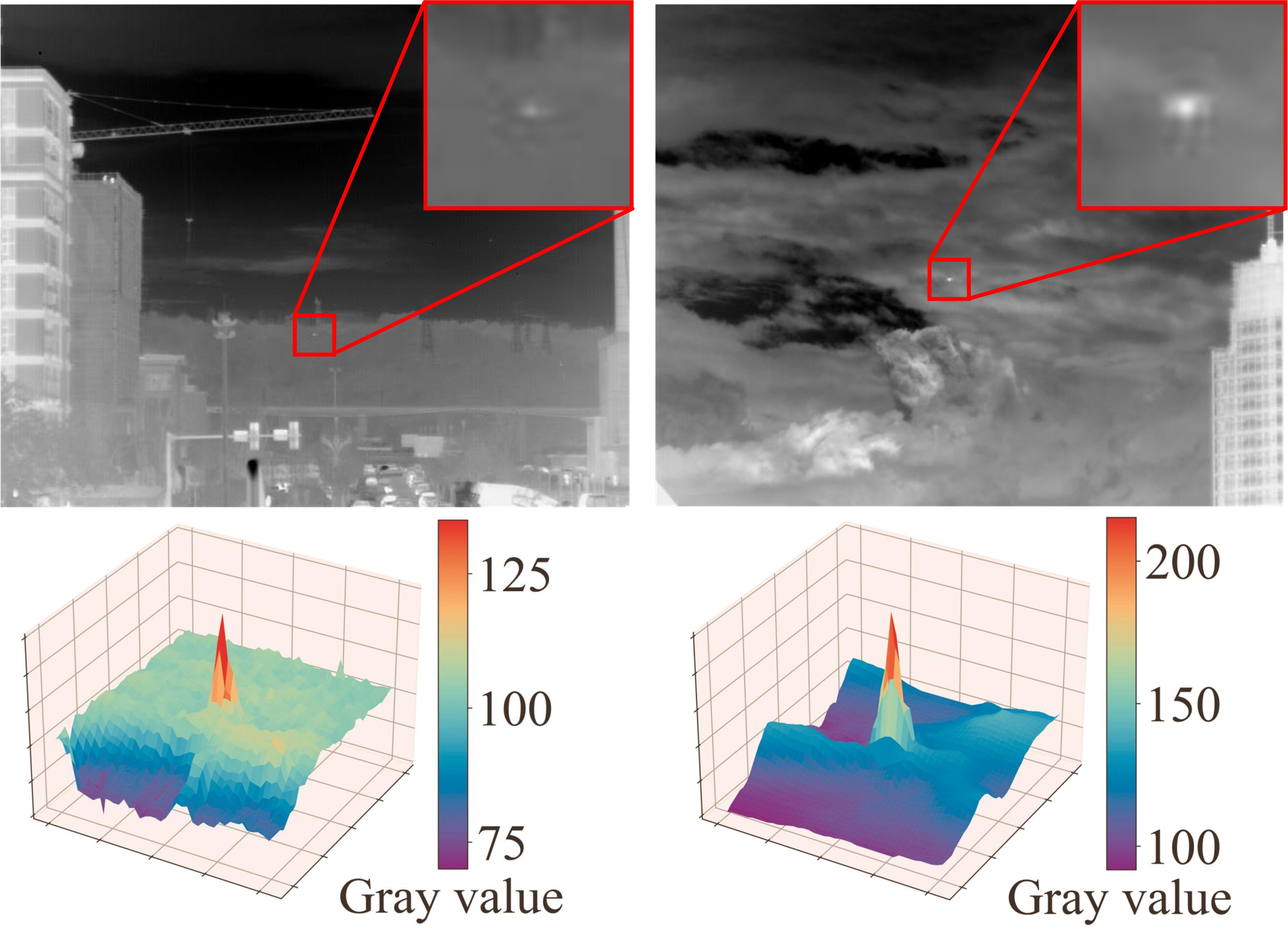}
	\caption{Grayscale 3D view of IRST.}
\end{figure}
IRSTDS techniques can be broadly divided into model-driven traditional methods and data-driven deep learning (DL)-based methods. Traditional model-driven approaches \cite{8,9,12,13,18,19} require manual parameter adjustment based on prior knowledge, making them less adaptable to the diverse and complex scenarios encountered in IRSTDS, resulting in poor robustness \cite{20}. In contrast, data-driven DL-based methods leverage extensive and varied IRSTDS data, allowing for autonomous parameter updates through gradient descent on the loss function, leading to more robust performance. CNN-based IRSTDS methods are primarily divided into detection-based methods \cite{21,23,oscar,eflnet} and segmentation-based methods \cite{25,26,27,28}. While most research improves IRSTDS through novel network architectures, we enhance performance by revisiting the convolutional module itself. As shown in Fig. 1, our analysis of the 3D grayscale distribution of infrared small targets (IRST) reveals that they exhibit Gaussian characteristics. Therefore, we propose a new plug-and-play pinwheel-shaped convolution (PConv) module, which better aligns with the imaging characteristics of IRST. Compared to standard convolution (Conv), PConv enhances bottom-layer feature extraction and expands the receptive field.

\begin{figure}[t]
	\centering
	\includegraphics[width=1.0\linewidth]{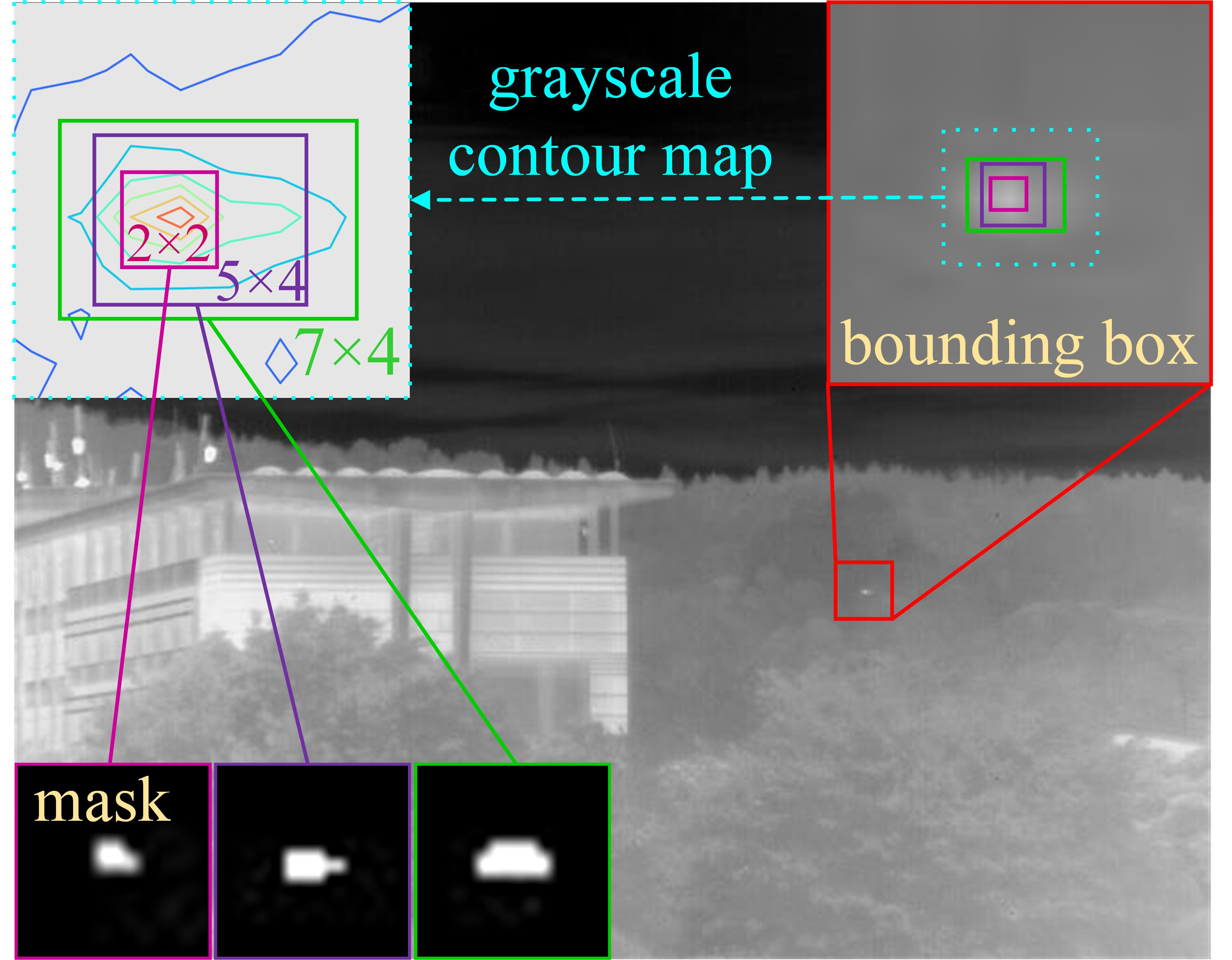}
	\caption{Visualization of BBox and mask label errors.}
\end{figure} 
Due to the dim-small characteristics of IRST and the subjectivity of manual labeling, as shown in Fig. 2, both bounding box (BBox) and mask labels exhibit significant IoU fluctuation errors. Distance IoU (DIoU) loss and complete IoU (CIoU) loss for BBox labels \cite{31}, and scale and location-sensitive (SLS) loss \cite{28} for mask labels, build on IoU loss by emphasizing positional information. However, they overlook IoU fluctuations and varying sensitivity to scale and location across different target sizes. To address this, we introduce a dynamic adjustment mechanism for the scale (S) and location (L) loss coefficients based on target size, integrating it into both DIoU and SLS losses. This aims to improve the regression performance of DL-based methods across varying target scales, leading to better detection performance.

Existing real shot IRSTDS datasets \cite{25,26} have a low proportion of small targets, simple backgrounds, and small data scales, which hinder detector performance in complex real-world scenarios. To address these limitations, we developed a new single-frame IRSTDS dataset, SIRST-UAVB, capturing unmanned aerial vehicle (UAV) and birds.

Our contributions can be summarized as follows:
\begin{itemize}
	\item Based on the Gaussian spatial distribution of IRST, we propose a novel plug-and-play convolutional module, PConv, which enhances CNNs' ability to analyze bottom-layer features of IRST.
	\item We introduce SD loss, which dynamically adjusts the impact coefficients of Scale loss and Location loss, improving the neural network’s regression capability and detection performance across targets of varying scales.
	\item We construct SIRST-UAVB, the largest publicly available dataset for real IRSTDS, encompassing comprehensive spatial domain challenges.
	\item We apply PConv and SD Loss to both BBox and mask label formats in IRSTDS methods, validating their effectiveness and generalization across public datasets and our own. Experimental results demonstrate significant and consistent performance improvements.
\end{itemize}
\begin{figure*}[!htbp]
	\centering
	\includegraphics[width=\linewidth]{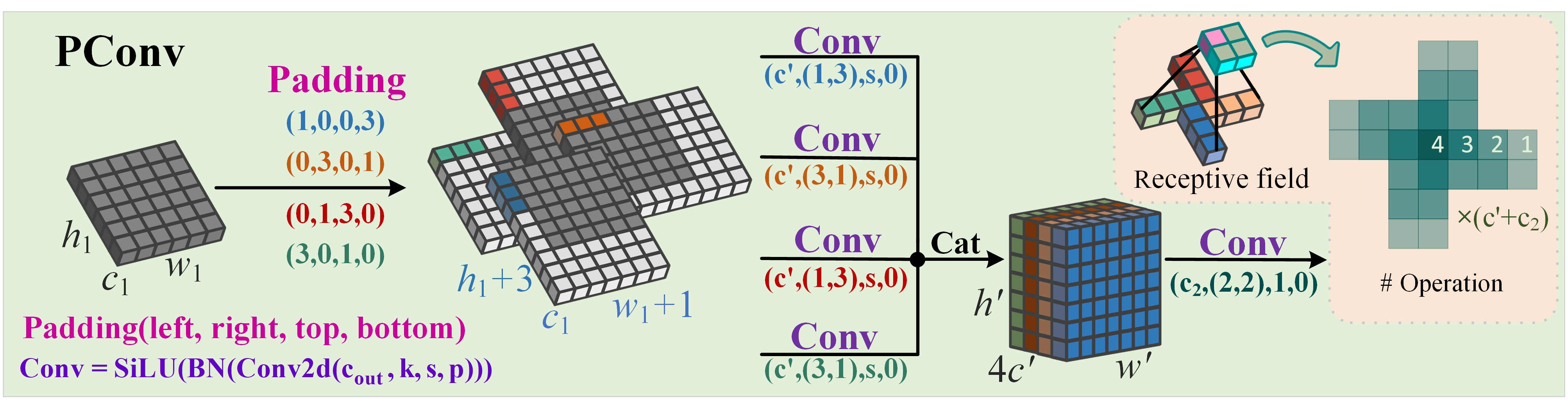}
	\caption{Architecture of the pinwheel-shaped convolutional module. Best viewed in color.}
\end{figure*}  

\section{Related Work}
\subsection{IRST Detection and Segmentation Networks}
Data-driven deep learning (DL) methods, which automatically learn multi-layered target features, have become the mainstream approach with the growth of DL techniques and IRSTDS data. In CNN-based IRST detection, Dai et al. proposed OSCAR \cite{oscar}, a one-stage cascade refinement network, and Yang et al. introduced EFLNet \cite{eflnet} to address target-background imbalance. For CNN-based IRST segmentation, the ACM module \cite{25} embeds low-level details into high-level features, while DNANet \cite{27} exploits contextual information through fusion. Liu et al. designed MSHNet \cite{28} with a Multi-Scale Head for U-Net. These CNN-based networks focus on building feature extraction networks but overlook the potential to enhance IRSTDS performance through the convolutional module. ISNet \cite{26}, which incorporates deformable convolutions, improves performance but demands increased training time and network parameters. 

Instead of focusing on network architecture design, our goal is to enhance the model's ability to extract bottom-layer features by introducing a convolutional module that aligns more closely with the IRST Gaussian distribution.

\subsection{Loss Functions for IRSTDS}
The loss function guides the model in minimizing the discrepancy between predicted and true labels, directly influencing the performance of DL-based models. BBox-based losses, such as generalised IoU (GIoU) \cite{33} and CIoU \cite{31}, combine overlap area, central point distance, and aspect ratio to enhance model convergence and accuracy. However, CIoU does not account for IoU fluctuation errors or the sensitivity of small-scale targets, particularly weak ones. The normalized Gaussian wasserstein distance (NWD) \cite{nwd} and scale adaptive fitness (SAFit) \cite{safit} were introduced to address these issues, but their reliance on exponential operations introduces complexity and instability. Mask-based IoU loss and Dice loss \cite{34} fail to account for target scale and location information. To address this, Liu et al. \cite{28} proposed SLS loss. However, SLS also overlooks the varying contributions of IoU and central point location for targets of different scales under mask labeling, which limits model performance.

To address these issues, we propose Scale-based Dynamic (SD) loss, which varies the influence of SLoss and Lloss based on target size. This simple approach reduces IoU fluctuation and leads to stable performance improvements.

\subsection{Datasets for IRSTDS}
Existing IRSTDS datasets are limited in size, diversity, and detection difficulty, which hinders research progress. The SIRST \cite{25} contains high-quality images but lacks complexity, with only 427 images. Similarly, the IRSTD-1K \cite{26} offers higher quality but includes fewer small targets and simpler challenges. The NUDT-SIRST \cite{27} generated simulation targets, limiting its applicability to real-world scenarios. Hence, we developed the SIRST-UAVB dataset, the largest publicly available dataset with weak targets, complex backgrounds, and challenging single-frame IRSTDS data.

\section{Methodology}
\subsection{Pinwheel-shaped Convolution}
The architecture of the PConv module is shown in Fig. 3. Unlike the Conv, PConv uses asymmetry padding to create horizontal and vertical convolution kernels for different regions of the image. The kernels diffuse outward, with ${{h}_{1}}$, ${{w}_{1}}$, ${{c}_{1}}$ representing the height, width, and channel size of the input tensor ${{X}^{({{h}_{1}},{{w}_{1}},{{c}_{1}})}}$. 
To enhance training stability and speed, we apply batch normalization (BN) and sigmoid linear unit (SiLU) after each convolution. The first layer of PConv performs parallel convolutions as follows:
\begin{equation}
	\begin{split}
		{{X}_{1}}^{({h}',{w}',{c}')}=SiLU(BN(X_{P(1,0,0,3)}^{({{h}_{1}},{{w}_{1}},{{c}_{1}})}\otimes {{W}_{1}}^{(1,3,{c}')})), \\ 
		{{X}_{2}}^{({h}',{w}',{c}')}=SiLU(BN(X_{P(0,3,0,1)}^{({{h}_{1}},{{w}_{1}},{{c}_{1}})}\otimes {{W}_{2}}^{(3,1,{c}')})), \\ 
		{{X}_{3}}^{({h}',{w}',{c}')}=SiLU(BN(X_{P(0,1,3,0)}^{({{h}_{1}},{{w}_{1}},{{c}_{1}})}\otimes {{W}_{3}}^{(1,3,{c}')})), \\ 
		{{X}_{4}}^{({h}',{w}',{c}')}=SiLU(BN(X_{P(3,0,1,0)}^{({{h}_{1}},{{w}_{1}},{{c}_{1}})}\otimes {{W}_{4}}^{(3,1,{c}')})). 
	\end{split}
\end{equation}
where $\otimes $ is the convolution operator,  ${{W}_{1}}^{(1,3,{c}')}$ is a $1\times3$ convolution kernel with an output channel of $c'$. The padding parameters $P(1,0,0,3)$ denote the number of padding pixels in the left, right, top, and bottom directions, respectively. The output feature map's height (${h}'$), width (${w}'$), and channel count (${c}'$) after the first layer of interleaved convolution are related to the input feature map as follows:
\begin{equation}
	{h}'=\frac{{{h}_{1}}}{s}+1, \quad {w}'=\frac{{{w}_{1}}}{s}+1, \quad {c}'=\frac{{{c}_{2}}}{4},
\end{equation}
where ${{c}_{2}}$ is the number of channels of the final output feature map from the PConv module, and $s$ is the convolution stride. The results of the first layer's interleaved convolution are concatenated ($Cat(.,.)$), and the output is computed as: 
\begin{equation}
	{{X}'}^{({h}',{w}',4{c}')}=Cat({{X}_{1}}^{({h}',{w}',{c}')},...,{{X}_{4}}^{({h}',{w}',{c}')}).
\end{equation}

Finally, the concatenated tensor is normalized by a convolution kernel ${{W}^{(2,2,{{c}{2}})}}$, without padding. The output feature map's height and width are adjusted to the preset values $h_2$ and $w_2$, making PConv interchangeable with Conv layers and serving as a channel-attention mechanism that analyzes the contribution of different convolutional orientations. The final output ${Y}^{({{h}_{2}},{{w}_{2}},{{c}_{2}})}$ are computed as:
\begin{equation}
	{{h}_{2}}={h}'-1=\frac{{{h}_{1}}}{s}, \quad {{w}_{2}}={w}'-1=\frac{{{w}_{1}}}{s},
\end{equation}
\begin{equation}
	{{Y}^{({{h}_{2}},{{w}_{2}},{{c}_{2}})}}=SiLU(BN({{X}'}^{({h}',{w}',{4c}')}\otimes {{W}^{(2,2,{{c}_{2}})}})).
\end{equation}

The receptive field's effectiveness diminishes outward, resembling a Gaussian distribution \cite{37}. Additionally, the smaller the target, the more concentrated its features become, which highlights the importance of central features. The upper right of Fig. 3 displays the receptive field of PConv (k=3) is 25, with the number of convolutions decreasing from the center outward, resembling a Gaussian distribution. Notably, PConv utilizes grouped convolution \cite{gconv}, significantly increasing the receptive field while minimizing the number of parameters. The number of parameters for Conv are calculated as:
\begin{equation}
	Conv_{params}={{c}_{2}}\times {{c}_{1}}\times k,(bias=False),	
\end{equation}
where $k$ is the convolution kernel size. If the number of output channels $(c_2)$ equals the input channels $(c_1)$, the  Conv's parameters are $9{{c_1}^{2}}$, and our PConv's parameters are calculated as follows:
\begin{equation}
	\begin{split}
		PConv_{params}=&4\times (({{c}_{2}}/4)\times{{c}_{1}}\times3\times1)\\
		&+4{{c}_{2}}{{c}_{1}}=7{{c}_{2}}{{c}_{1}=7{c_1}^2}
	\end{split}
\end{equation}
representing a 22.2\% reduction compared to $9{{c}_{1}}^{2}$ for $3\times3$ Conv, while increasing the receptive field by 177\%. However, since PConv is used to extract bottom-layer features in IRST, it replaces the first two Conv layers, such as in the YOLO series, where ${{c}_{2}}=4{{c}_{1}}$. In this context, Conv requires $36{{c}_{1}}^{2}$ parameters, while PConv requires $72{{c}_{1}}^{2}$. Thus, PConv(k=3) increases the receptive field by 178\% with only a 111\% increase in parameters compared to a $3\times3$ Conv. Similarly, PConv(k=4) increases the receptive field by 444\% with just a 122\% increase in parameters. This demonstrates that PConv achieves an efficient receptive field expansion with minimal parameter increase. 

\begin{figure}[t]
	\centering
	\includegraphics[width=\linewidth]{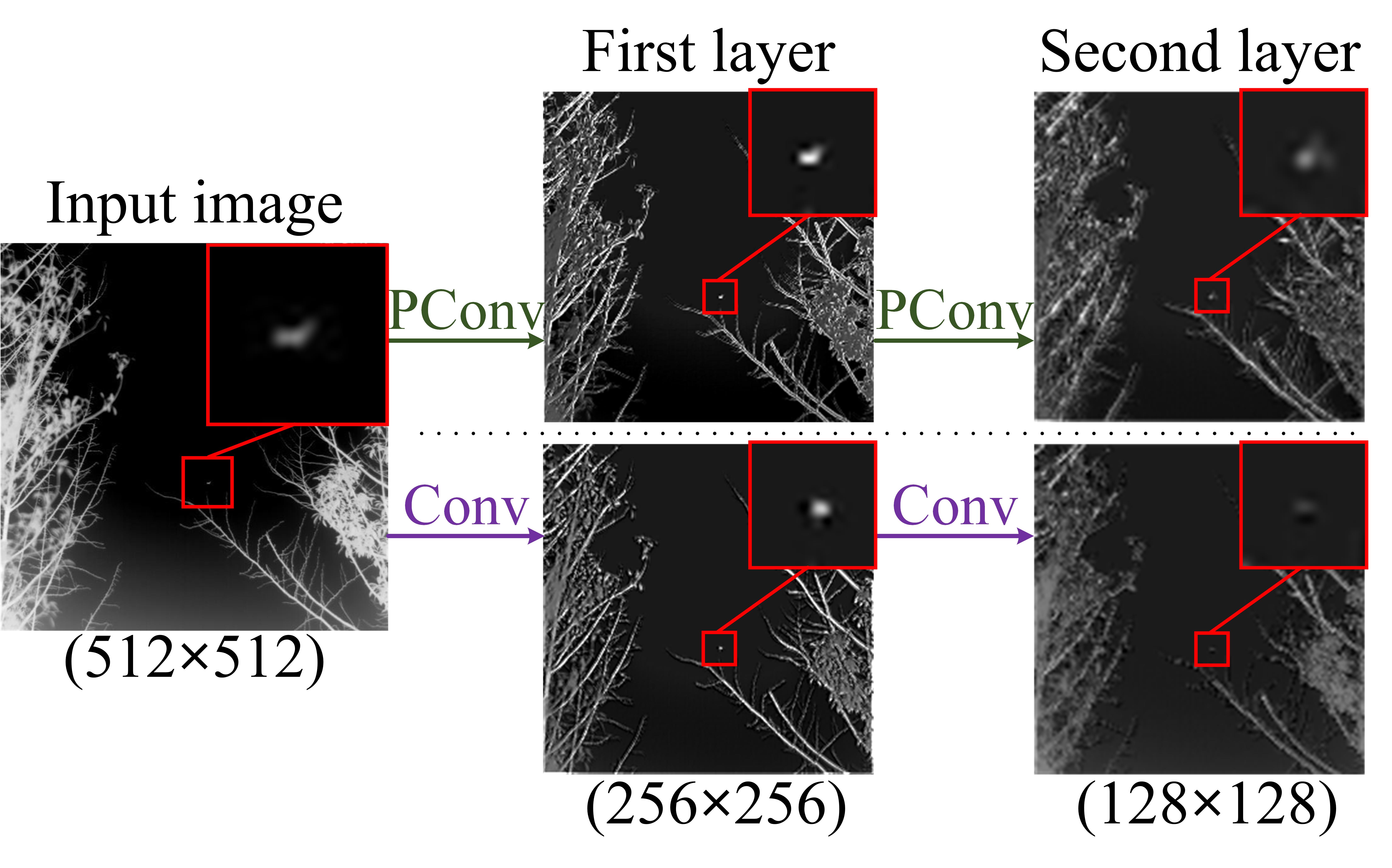}
	\caption{Visualization of PConv and Conv output results.}
\end{figure}
Additionally, we calculated the average value of multiple channels from the PConv and Conv output results to obtain the visual results shown in Fig. 4. These results demonstrate that PConv enhances the contrast between IRST targets and the background while suppressing clutter-like signals.

\subsection{Scale-based Dynamic Loss}
The upper right corner of Fig. 2 shows that the IoU-based loss (Sloss) fluctuates by up to 86\%. Smaller targets experience greater instability in IoU loss, negatively impacting model stability and regression.
However, we observed that regardless of BBox size, the centroid coordinates deviate by no more than 1 pixel from the target's center of gravity.
Therefore, we dynamically adjust the influence coefficients of Sloss and Lloss based on target scale, reducing the impact of label inaccuracies on loss function stability.
As shown in Fig. 5(a), smaller targets receive lower attention weights for Sloss with BBox labels. Mask labels can improve detection accuracy, especially for small or irregularly shaped targets. However, the fuzzy boundaries of IRST, as shown in the bottom left of Fig. 2, lead to a 62\% Sloss fluctuation. Smaller targets further increase this instability.
\begin{figure}[htbp]
	\centering
	\includegraphics[width=\linewidth]{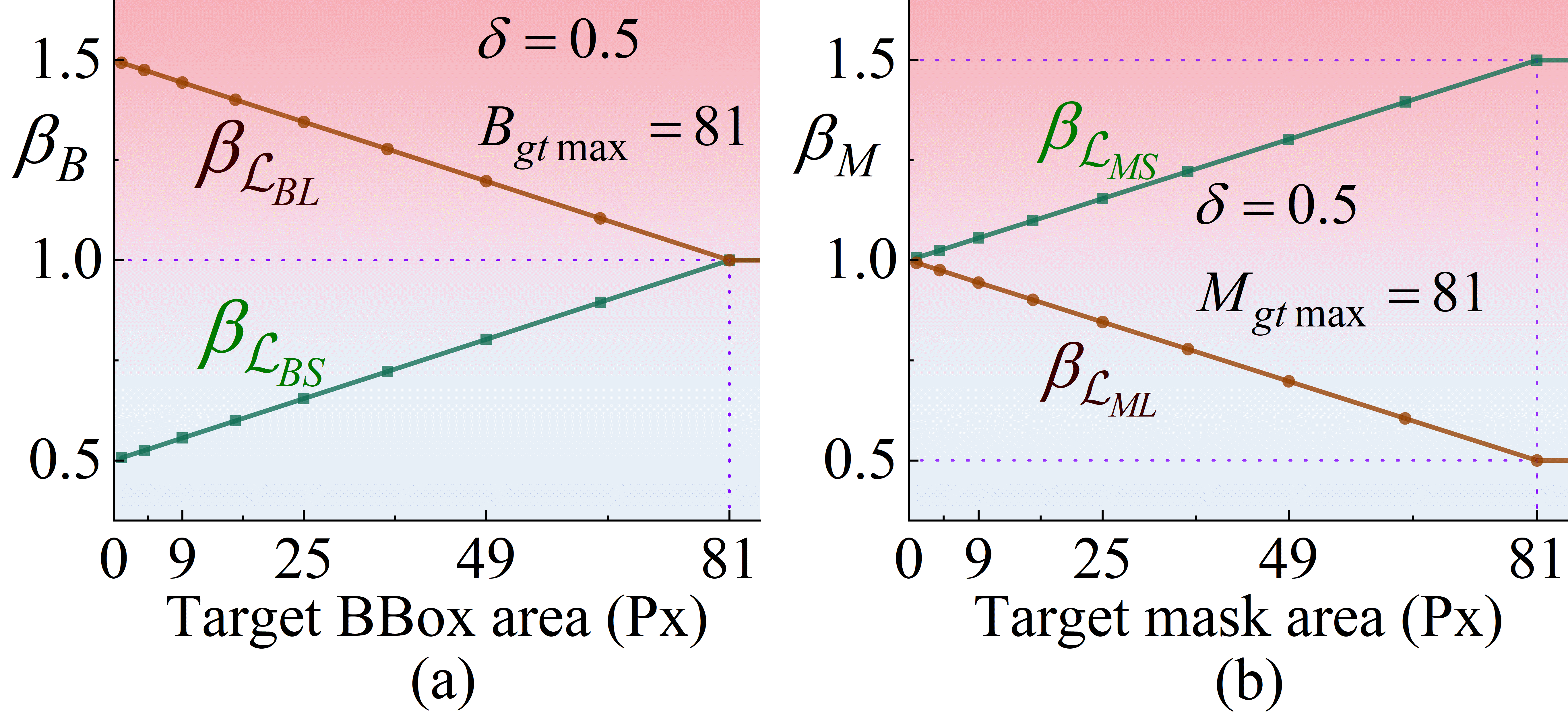}
	\caption{(a) The value of weight ${{\beta }_{B}}$ in Sloss and Lloss concerning the target BBox area. (b) The value of weight ${{\beta }_{M}}$ in Sloss and Lloss concerning the target mask area.}
\end{figure}
Additionally, Lloss for mask labels considers the average position of all objects in an image, making it difficult to converge when one object is missed, leading to more false alarms. 
Therefore, as shown in Fig. 5(b), we enhance the influence of Sloss for mask labels to ensure the model focuses more on Sloss.
\begin{equation}
	{{\mathcal{L}}_{BS}}=1-IoU+\alpha v, \quad {{\mathcal{L}}_{BL}}=\frac{{{\rho }^{2}}({{b}_{p}},{{b}_{gt}})}{{{c}^{2}}}.
\end{equation}
Here, $IoU$ represents the Intersection over the Union of the predicted and ground truth BBox, $\alpha v$ measures the aspect ratio consistency of the BBox, $\rho (.)$ is the Euclidean distance, ${{b}_{p}}$ and ${{b}_{gt}}$ are the centroids of the predicted BBox ${{B}_{p}}$ and target BBox ${{B}_{gt}}$, and $c$ is the diagonal length of two BBoxes.
\begin{figure*}[h]
	\centering
	\includegraphics[width=\linewidth]{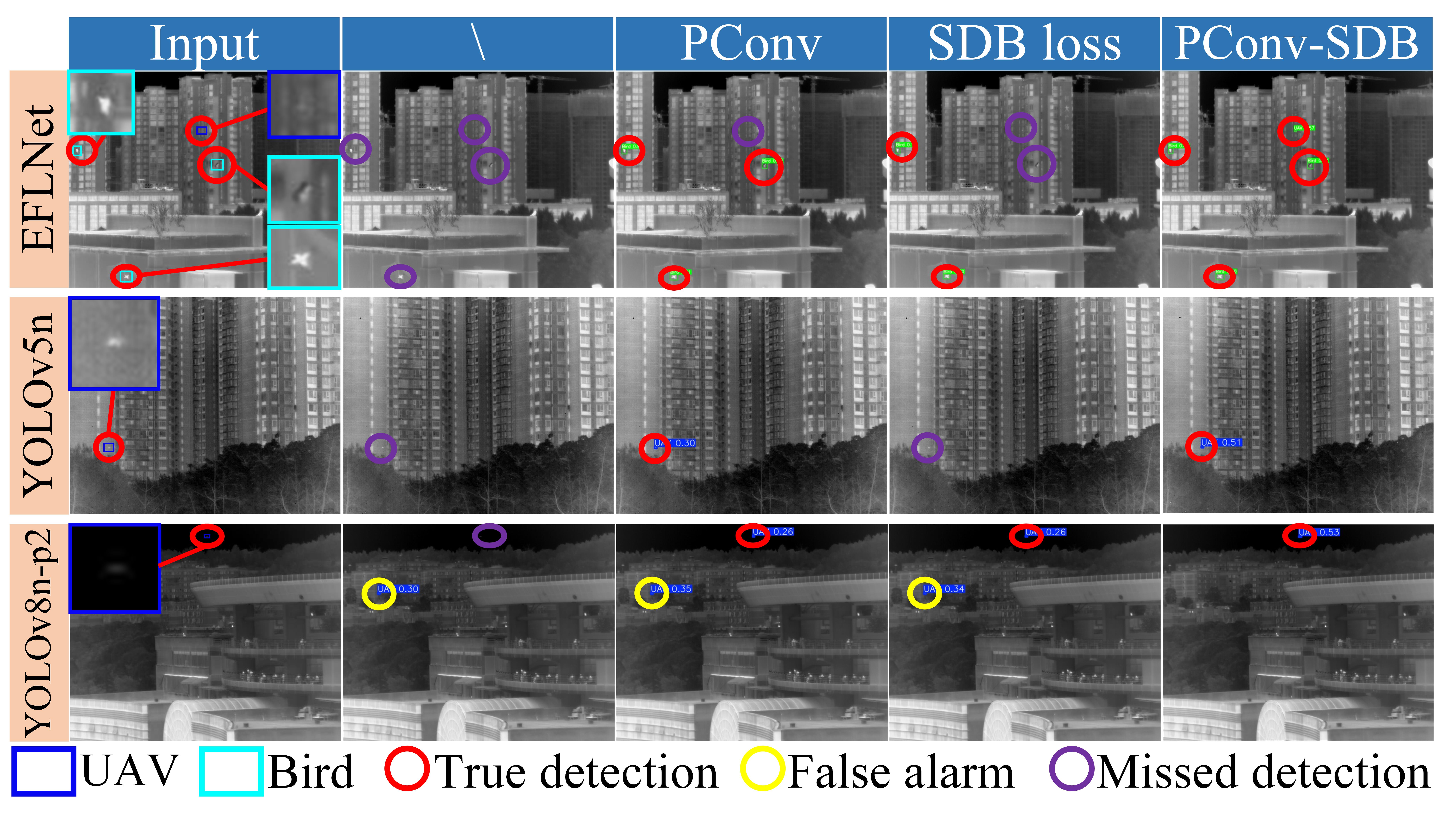}
	\caption{Result visualization of IRST detection models.}
\end{figure*} 

For the segmentation loss function of mask, we refer to SLS \cite{28} loss and define the mask scale loss (${{\mathcal{L}}_{MS}}$) and the mask location loss (${{\mathcal{L}}_{ML}}$) as follows:
\begin{equation}
	{{\mathcal{L}}_{MS}}=1-\omega \frac{\left| {{M}_{p}}\bigcap {{M}_{gt}} \right|}{\left| {{M}_{p}}\bigcup {{M}_{gt}} \right|}
\end{equation}
\begin{equation}
	{{\mathcal{L}}_{ML}}=1-\frac{\min ({{d}_{p}},{{d}_{gt}})}{\max ({{d}_{p}},{{d}_{gt}})}+\frac{4}{{{\pi }^{2}}}{{({{\theta }_{p}}-{{\theta }_{gt}})}^{2}}
\end{equation}
where ${{M}_{p}}$ and ${{M}_{gt}}$ are the set of predicted pixels and ground truth pixels of targets, ${{d}_{p}}$ and ${{d}_{gt}}$ are the distances of the predicted and target mean pixels from the origin in polar coordinates, and ${{\theta }_{p}}$ and ${{\theta }_{gt}}$ are the mean angles of the predicted and target pixels in polar coordinates, $\omega$ describes the difference between ${{M}_{p}}$ and ${{M}_{gt}}$.
When the model scales the image or subsamples the feature map, the target size changes. To determine the true target size, we calculate the ratio between the original image and the current feature map:
\begin{equation}
	{{R}_{OC}}=\frac{{{w}_{o}}\times {{h}_{o}}}{{{w}_{c}}\times {{h}_{c}}},
\end{equation}
where ${{w}_{o}}$, ${{h}_{o}}$ are the width and height of the original image, and ${{w}_{c}}$, ${{h}_{c}}$ are those of the current feature map. The influence coefficient of BBox $({{\beta }_{B}})$ and mask $({{\beta }_{M}})$ are calculated as:
\begin{equation}
	{{\beta }_{B}}=min(\frac{{{B}_{gt}}}{{{B}_{gt\max }}}\times {{R}_{OC}}\times \delta ,\delta),
\end{equation}
\begin{equation}
	{{\beta }_{M}}=min(\frac{{{M}_{gt}}}{{{M}_{gt\max }}}\times {{R}_{OC}}\times \delta ,\delta),
\end{equation}
where ${{B}_{gt\max }}={{M}_{gt\max }}=81$ is the maximum size of IRST as defined by the Society of Photo-Optical Instrumentation Engineers \cite{36}. The impact coefficient of the loss is based on the current target box's area, and its range is restricted to $\delta$, which is adjustable. The final Scale-based Dynamic Loss for the BBox (SDB loss) is given by:
\begin{equation}
	{{\beta }_{{{\mathcal{L}}_{BS}}}}=1-\delta +{{\beta }_{B}}, \quad {{\beta }_{{{\mathcal{L}}_{BL}}}}=1+\delta -{{\beta }_{B}},
\end{equation}
\begin{equation}
	{{\mathcal{L}}_{SDB}}={{\beta }_{{{\mathcal{L}}_{BS}}}}\times {{\mathcal{L}}_{BS}}+{{\beta }_{{{\mathcal{L}}_{BL}}}}\times {{\mathcal{L}}_{BL}}.
\end{equation}
where ${{\beta }_{{{\mathcal{L}}_{BS}}}}$ and ${{\beta }_{{{\mathcal{L}}_{BL}}}}$ are the impact factors for ${{\mathcal{L}}_{BS}}$ and ${{\mathcal{L}}_{BL}}$, respectively. When the target BBox area is larger than 81, ${{\mathcal{L}}_{SDB}}$ degenerates into CIoU loss.
The Scale-based Dynamic Loss for Mask (SDM loss) can be calculated as:
\begin{equation}
	{{\beta }_{{{\mathcal{L}}_{MS}}}}=1+{{\beta }_{M}}, \quad {{\beta }_{{{\mathcal{L}}_{ML}}}}=1-{{\beta }_{M}},  
\end{equation}
\begin{equation}
	{{\mathcal{L}}_{SDM}}={{\beta }_{{{\mathcal{L}}_{MS}}}}\times {{\mathcal{L}}_{MS}}+{{\beta }_{{{\mathcal{L}}_{ML}}}}\times {{\mathcal{L}}_{ML}}.
\end{equation}
where ${{\beta }_{{{\mathcal{L}}_{MS}}}}$ and ${{\beta }_{{{\mathcal{L}}_{ML}}}}$ of ${{\mathcal{L}}_{MS}}$ are the impact factors for ${{\mathcal{L}}_{MS}}$ and ${{\mathcal{L}}_{ML}}$, respectively.  

\subsection{SIRST-UAVB Dataset}
We created a benchmark called SIRST-UAVB, consisting of 3,000 infrared images targeting UAVs and birds, collected over a year across various seasons, weather conditions, and complex backgrounds. The dataset presents challenges like diverse target orientations, scales, and occlusions, with a high proportion of small targets, many of which are nearly invisible to the naked eye. We manually annotated the targets based on trajectories, ensuring accuracy through repeated checks. The dataset includes 1,742 bird and 2,955 UAV BBox labels, but due to the difficulty of accurately labeling faint bird targets, we excluded them from the mask annotations. Overall, SIRST-UAVB is well-suited for DL-based detection in complex real-world scenarios.

\begin{table*}[htbp]
	\normalsize
	\centering
	\begin{tabular}{c@{\hskip 0.1in}|@{\hskip 0.1in}c@{\hskip 0.1in}c@{\hskip 0.1in}c@{\hskip 0.1in}|@{\hskip 0.1in}c@{\hskip 0.1in}c@{\hskip 0.1in}c@{\hskip 0.1in}|c|@{\hskip 0.1in}c@{\hskip 0.1in}c@{\hskip 0.1in}c@{\hskip 0.1in}|@{\hskip 0.1in}c@{\hskip 0.1in}c@{\hskip 0.1in}c}
		\toprule
		& \multicolumn{7}{c@{\hskip 0.1in}|@{\hskip 0.1in}}{YOLOv8n-p2 detection} & \multicolumn{6}{c}{MSHNet segmentation} \\ \midrule
		\multirow{2}{*}{\makecell{Convolution\\module}}& \multicolumn{3}{c@{\hskip 0.1in}|@{\hskip 0.1in}}{IRSTD-1K} & \multicolumn{3}{c|}{SIRST-UAVB} &  & \multicolumn{3}{c@{\hskip 0.1in}|@{\hskip 0.1in}}{IRSTD-1K} & \multicolumn{3}{c}{SIRST-UAVB} \\ 
		& $P$ & $R$ & $mAP50$ & $P$ & $R$ & $mAP50$ & $Params$ & $IoU$ &  ${{P}_{d}}$ & ${{F}_{a}}\downarrow$  & $IoU$ & ${{P}_{d}}$ & ${{F}_{a}}\downarrow$ \\ \midrule
		Conv       & 89.0 & 83.7 & 87.4  & 90.8  & 89.4  & 93.2 &  \underline{2.786} & 66.82 & 93.54 & 14.20 & 23.93 & \underline{72.66} & 11.37    \\  
		GConv  & 89.9 & 83.0   & 87.6  & 87.5  & 85.5  & 89.7 & \textbf{2.782} & 67.12  & 92.71  & 16.48  & 24.07 &  72.31  & 12.72 \\
		DSConv & 87.8 & 83.7 &  87.5 & 89.9  & 89.1   & 92.6 & \textbf{2.782} & 66.54 & 93.79 & 15.25 &  23.39 &  71.58  &  13.19 \\
		DRConv & 90.0 & 83.1 & 87.6 & 91.5 & 88.7 & 92.8 & 2.789 & 66.42 & 93.88 & 13.17 & 22.70 & 71.62& 13.03  \\ 
		LSKConv&88.7 & 84.6 & 88.3 & 91.9 & 90.2 & 93.2 & 2.791 & 67.13 & 94.06 & 17.24 & 23.74 & 72.53 & 14.26 \\
		DConv  &90.4&80.1&86.6&88.5&77.8&84.0 & \underline{2.786} & 66.73 & 91.25 &16.38 &20.50&69.25&15.36 \\
		MixConv& 88.3 & \underline{85.9} & 88.2 &  91.3 & 89.0 & 91.9 & 2.802 & 63.44 & 90.23 & 30.44 & 21.90 & 72.59 & \underline{11.55} \\ 
		AKConv & \underline{90.5} & 82.2 & 86.9 & 89.6 & 89.2 & 92.7 & 2.791 & 64.32 & 91.34 & 17.45 & 21.97 & 71.56 & 12.19 \\  \midrule
		PConv(3,3) & 88.0 & 85.0 & \underline{89.2}  & 92.1  & \underline{90.0}    & \underline{93.4} & 2.787 & 67.06 & \textbf{95.58} & 16.17 & \underline{24.49} & 72.40 & 11.65 \\
		PConv(4,3) & \textbf{90.7} & 83.9 & \textbf{89.9} & \underline{92.6} & \textbf{90.5} & \textbf{93.8} & 2.787 & \textbf{67.93} & \underline{94.22} & \textbf{9.19} & \textbf{24.66} & \textbf{73.70} & \textbf{10.83}  \\ 
		PConv(4,4)& 90.0  & \textbf{86.0} & 89.1 & \textbf{93.7}  & 89.6  & \underline{93.4} & 2.787 & \underline{67.45} & 92.20 & \underline{10.70} & 23.66 & 72.40 & 13.30    \\ 
		\bottomrule
	\end{tabular}
	\caption{We evaluate various convolution modules by replacing the first two standard layers in YOLOv8n-p2 (using CIoU loss) detection and MSHNet (using SLS loss) segmentation frameworks. PConv uses different "fanleaf" lengths (e.g., '4, 3' means the first PConv kernel is 4 and the second is 3). Detection results are evaluated by $P(\%)$, $R(\%)$, and $mAP50(\%)$. $Params(M)$ indicates the number of parameters. Segmentation results are evaluated by $IoU(\%)$, ${{P}{d}}(\%)$, and ${{F}{a}}(10^{-6})\downarrow$ (where $\downarrow$ means lower is better). Results are highlighted in bold for the best performance and underlined for the second-best performance.}
	\label{combined_table}
\end{table*}
%----------------------------------------------------------------------------------------------------------------
\section{Experiments}
\subsection{Experimental Settings}
\subsubsection{Datasets.} 
To evaluate the impact of PConv and SD loss across different hyperparameters on targets of varying scales, we used two datasets: IRSTD-1K \cite{26}, containing 1,000 real infrared images with an average larger target scale and a resolution of $512\times512$ pixels, and our SIRST-UAVB, which features smaller targets. Both datasets were split into training and testing sets with a 4:1 ratio.
\subsubsection{Evaluation Metrics for Bounding Box Label.} Since our SIRST-UAVB has both UAV and Bird targets, we evaluate the accuracy of model detection and classification using precision ($P$) and recall ($R$). $P$ denotes the proportion of samples predicted to be true that are indeed true, and $R$ denotes the proportion of true samples that are detected. 
\begin{equation}
	\begin{split}
		P=\frac{TP}{TP+FP}&, \quad R=\frac{TP}{TP+FN}, \\
		mAP50&=\frac{1}{C}\sum\limits_{i=1}^{C}{A{{P}_{i,50}}}.
	\end{split}
\end{equation}
where $TP$ is the true positive (The $IoU$ is greater than the threshold of 0.45) in the predicted positive samples, $FP$ is the false positive in the predicted positive samples, $FN$ is the false negative in the predicted negative samples. We used Mean Average Precision 50 ($mAP50$) to assess the metrics for detecting the accuracy of the target location.
\subsubsection{Evaluation Metrics for Mask Label.} For mask labels, we use $IoU$ to evaluate the model's pixel-level shape description capability, and the False alarm (${{F}_{a}}$) rate and the probability of detection (${{P}_{d}}$) to evaluate localization performance. ${{P}_{false}}$ is the false positive pixels, ${{P}_{all}}$ is the total pixels, ${{T}_{pred}}$ is the correctly predicted targets, and ${{T}_{all}}$ is the total targets. ${{F}_{a}}$and ${{P}_{d}}$ are calculated as:
\begin{equation}
	IoU=\frac{\left| {{A}_{p}}\bigcap {{A}_{gt}} \right|}{\left| {{A}_{p}}\bigcup {{A}_{gt}} \right|}, \quad
	{{F}_{a}}=\frac{{{P}_{false}}}{{{P}_{all}}}, \quad
	{{P}_{d}}=\frac{{{T}_{pred}}}{{{T}_{all}}}.
\end{equation}
\begin{figure*}[h]
	\centering
	\includegraphics[width=\linewidth]{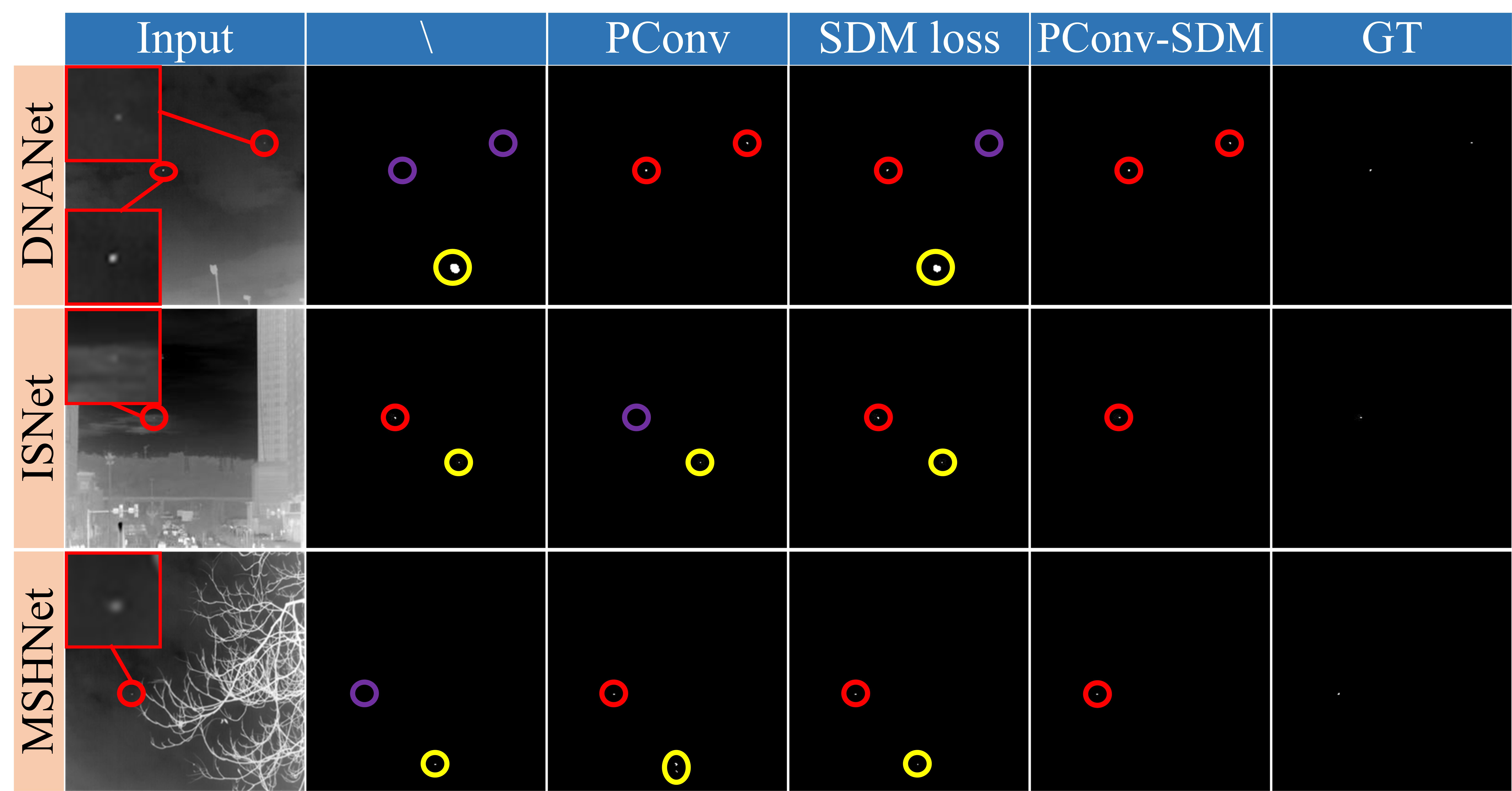}
	\caption{Result visualization of IRST segmentation models.}
\end{figure*} 

\subsubsection{Implementation Detail.}
We conducted ablation experiments on IRST detection and segmentation models, using the PyTorch framework on RTX3090 GPUs. For detection models, the input image size was set to 640, batch size to 64, epochs to 700, patience to 70, and learning rate to 0.01. For segmentation models, the input image size was set to 256, batch size to 4, epochs to 400, and learning rate to 0.05.

\subsection{Comparison with Other Methods} 
\subsubsection{Comparison of Convolution Module.}
As shown in Table 1, we compared PConv with various convolutional modules. Group convolution (GConv) \cite{gconv} and depthwise separable convolution (DSConv) \cite{sdconv} focus on reducing parameters, while dynamic region-aware convolution (DRConv) \cite{drconv}, large selective kernel convolution (LSKConv) \cite{lskconv}, dilated convolution (DConv) \cite{dconv}, MixConv \cite{mixconv}, and AKConv \cite{akconv} aim to expand the receptive field.

In the YOLOv8n-p2 detection model, most alternative modules did not consistently enhance performance, except for MixConv. However, MixConv required more parameters but still did not outperform our proposed PConv. For the IRSTD-1K dataset, the YOLOv8n-p2 model with PConv(4,4) achieved the best overall performance, as indicated by the highest mean evaluation metrics. However, the configuration with PConv(4,3) demonstrated the most balanced improvement, achieving the most top evaluation metrics. For the SIRST-UAVB dataset, the PConv(4,3) provided the best and most balanced performance enhancement. This indicates that for larger targets in the IRSTD-1K dataset, a larger PConv kernel length is beneficial, while for smaller targets in the SIRST-UAVB dataset, increasing the PConv kernel length does not yield additional performance gains.
In the MSHNet segmentation model, PConv significantly outperformed other convolution modules. The results indicate that a PConv kernel length of 4 in the first layer provides a more effective receptive field, crucial for capturing the features of small targets. As the feature map and target size reduce during downsampling, a kernel length of 3 is sufficient for subsequent layers, reducing computational overhead while maintaining performance.

These experiments clearly demonstrate that, compared to other convolution modules, PConv stands out due to its design, which aligns with the Gaussian distribution characteristics of IRST gray and effectively expands the convolutional receptive field. This enhances the network's ability to extract bottom-layer features for IRST with only a negligible increase in parameters.

\begin{table}[htbp]
	\normalsize
	\centering
	{
		\begin{tabular}{c@{\hskip 0.05in}|c@{\hskip 0.1in}c@{\hskip 0.05in}c|c@{\hskip 0.1in}c@{\hskip 0.05in}c}
			\toprule
			& \multicolumn{3}{c|}{IRSTD-1K} & \multicolumn{3}{c}{SIRST-UAVB} \\ %\cmidrule(lr){2-4} \cmidrule(lr){5-7}
			Loss & $P$ & $R$ & $mAP50$  & $P$ & $R$ & $mAP50$ \\ \midrule
			CIoU & 89.0 & 83.7 & 87.4 & 90.8 & \underline{89.4} & 93.2  \\
			DIoU & 90.1 & 83.6 & 87.4 & 91.3 & 88.4 & 93.6 \\
			GIoU & 89.5 & 83.1 & 88.1 & 91.5 & 88 & 92.5 \\
			IoU  & 89.7 & 83.9 & 87.7 & 92.2 & 89.2 & \textbf{94} \\
			NWD  & 89.3  & \underline{84.2}  & 88.5  & 92.4  & 88.9  & 93.7   \\
			SAFit & 88.3  & 83.7  & 88.0  & \textbf{93.6}  & 89.1  & 93.5  \\  \midrule
			SDB(0.3) & \textbf{91.7} & \textbf{85.4} & \underline{88.6} & 91.8 & \textbf{89.5} & \underline{93.9}    \\ 
			SDB(0.5) & \underline{90.7} & 84.1 & 88.1 & 92.7 & \textbf{89.5} & 93.8   \\ 
			SDB(0.7) & 90.2 & 83.7 & \textbf{88.8} & \underline{93.0} & 89.2 & \underline{93.9}      \\ \bottomrule
	\end{tabular}}
	\caption{Comparative experiments are conducted on YOLOv8n-p2 using various BBox loss and our SDB$(\delta)$ loss.}
\end{table}
\subsubsection{Comparison of Loss Functions.} 
Tables 2 and 3 summarize the performance of various loss functions for IRST detection and segmentation.
In Table 2, we compared several BBox-based losses, including CIoU, DIoU \cite{31}, GIoU \cite{33}, IoU, NWD \cite{nwd}, SAFit \cite{safit}, and our proposed SDB$(\delta)$ loss. While SAFit performed well on the SIRST-UAVB dataset, its performance significantly dropped on the IRSTD-1K dataset. In contrast, our SDB loss provided consistent and balanced improvement across both datasets, which is important for real-world applications with varying target sizes and distributions. Moreover, NWD and SAFit losses involve exponential operations, increasing computational complexity and time, whereas SDB is simpler and more efficient.
\begin{table}[htbp]
	\normalsize
	\centering
	{
		\begin{tabular}{c@{\hskip 0.05in}|@{\hskip 0.05in}c@{\hskip 0.1in}c@{\hskip 0.1in}c@{\hskip 0.05in}|@{\hskip 0.05in}c@{\hskip 0.1in}c@{\hskip 0.1in}c}
			\toprule
			& \multicolumn{3}{c@{\hskip 0.05in}|@{\hskip 0.05in}}{IRSTD-1K} & \multicolumn{3}{c}{SIRST-UAVB} \\ %\cmidrule(lr){2-4} \cmidrule(lr){5-7}
			Loss    & $IoU$ &  ${{P}_{d}}$ & ${{F}_{a}}\downarrow$  & $IoU$ &  ${{P}_{d}}$ & ${{F}_{a}}\downarrow$ \\ \midrule
			SLS  & 66.82 & \textbf{93.54} & 14.20 & 23.93 & 72.66 & 11.37       \\  
			Dice & 65.16 & 92.87 & 13.13 & 21.91 & 74.82 & 10.92 \\
			IoU & 65.89 & 92.45 & 15.82 & 22.59 & 71.59 & \underline{10.04} \\  \midrule
			SDM(0.3)& \underline{68.17} & \underline{93.20} & \textbf{7.06} & 24.36 & \textbf{75.52} & 10.23       \\ 
			SDM(0.5) & \textbf{68.49} & \textbf{93.54} & \underline{9.34} & \textbf{24.97} & \underline{75.00} & \textbf{9.97}     \\ 
			SDM(0.7) & 67.19 & \textbf{93.54} & 12.75 & 23.56 & 72.66 & 11.04      \\ \bottomrule
	\end{tabular}}
	\caption{Comparative experiments are conducted on MSHNet using various mask loss and our SDM$(\delta)$ loss.}
\end{table}

%----------------------------------------------------------------------------------------------
\begin{table*}[htbp]
	\normalsize
	\centering
	\begin{tabular}{c@{\hskip 0.1in}c|@{\hskip 0.05in}c@{\hskip 0.05in}|c@{\hskip 0.1in}c@{\hskip 0.05in}c@{\hskip 0.05in}|@{\hskip 0.05in}c@{\hskip 0.1in}c@{\hskip 0.05in}c@{\hskip 0.05in}|@{\hskip 0.05in}c@{\hskip 0.05in}|c@{\hskip 0.1in}c@{\hskip 0.1in}c@{\hskip 0.05in}|c@{\hskip 0.1in}c@{\hskip 0.1in}c}
		\toprule
		& & \multicolumn{7}{c@{\hskip 0.05in}|@{\hskip 0.05in}}{Detection} & \multicolumn{7}{c}{Segmentation} \\ \midrule
		& &  & \multicolumn{3}{c@{\hskip 0.05in}|@{\hskip 0.05in}}{IRSTD-1K} & \multicolumn{3}{@{\hskip 0.05in}c@{\hskip 0.05in}|@{\hskip 0.05in}}{SIRST-UAVB} & & \multicolumn{3}{c|}{IRSTD-1K} & \multicolumn{3}{c}{SIRST-UAVB} \\ 
		PConv & SD & Model &  $P$ & $R$ & $mAP50$ & $P$ & $R$ & $mAP50$ & Model & $IoU$ &  ${{P}_{d}}$ & ${{F}_{a}}\downarrow$  & $IoU$ &  ${{P}_{d}}$ & ${{F}_{a}}\downarrow$ \\ \midrule
		\XSolidBrush & \XSolidBrush & \multirow{4}{*}{\makecell{EFL\\Net}} &  87.2 & 79.7 & 82.1 & 88.8 & 83.4 & 89.3 & \multirow{4}{*}{\makecell{DNA\\Net}} & 60.94 & 91.16 & 15.11 & 23.96 & \textbf{76.56} & 13.91 \\
		\XSolidBrush & \CheckmarkBold & & \underline{87.4} & 80.2 & 82.9 & \underline{92.1} & \textbf{87.8} & \underline{91.2} & & 62.76 & 90.82 & 12.6 & 24.22 & 74.48 & \underline{10.38} \\
		\CheckmarkBold & \XSolidBrush & & 86.9 & \underline{80.6} & \underline{83.4} & 88.8 & 87.0 & 90.0 & & \underline{65.38} & \textbf{93.20} & 15.11 & 25.08 & 74.22 & 10.45 \\
		\CheckmarkBold & \CheckmarkBold & & \textbf{87.5} & \textbf{81.2} & \textbf{84.4} & \textbf{92.9} & \underline{87.1} & \textbf{91.5} & & \textbf{66.49} & \underline{92.18} & \textbf{7.06} & \textbf{25.73} & \underline{74.74} & \textbf{9.46} \\ \midrule
		\XSolidBrush & \XSolidBrush & \multirow{4}{*}{\makecell{YOLO\\v5n}} & 84.8 & 77.1 & 82.7 & 84.8 & 48.1 & 60.0 & \multirow{4}{*}{\makecell{IS\\Net}} & 63.37 & 92.51 & 13.05 & 23.11 & 73.61 & 14.22 \\
		\XSolidBrush & \CheckmarkBold &  & \underline{86.9}& 75.7 & \textbf{84.6} & \underline{87.2} & \textbf{49.6} & \underline{62.2} & & 63.76 & 92.92 & \underline{12.51} & 23.72 & \underline{74.08} & \underline{10.38} \\
		\CheckmarkBold & \XSolidBrush &  & 86.8 & \textbf{78.4}& \underline{83.4} & 85.3 & \underline{48.9} & 61.4 & & \textbf{65.22} & \underline{93.59} & 13.64 & \underline{24.18} & 73.24 & 13.45 \\
		\CheckmarkBold & \CheckmarkBold & & \textbf{87.3} & \underline{77.6} & \textbf{84.6} & \textbf{88.7} & \underline{48.9} & \textbf{63.6} & & \underline{65.19} & \textbf{93.98} & \textbf{9.86} & \textbf{24.80} & \textbf{74.88} & \textbf{10.22} \\ \midrule
		\XSolidBrush & \XSolidBrush &\multirow{4}{*}{\makecell{YOLO\\v8n-p2}} & 89.0 & 83.7 & 87.4 & 90.8 & 89.4 & 93.2 & \multirow{4}{*}{\makecell{MSH\\Net}} & 66.82 & 93.54 & 14.20 & 23.93 & 72.66 & 11.37 \\  
		\XSolidBrush & \CheckmarkBold & &  \textbf{90.7} &\textbf{84.1} & 88.1 & \textbf{92.7} & 89.5 & \underline{93.8} & & \textbf{68.49} & 93.54 & 9.34 & \textbf{24.97} & \textbf{75.00} & \textbf{9.97} \\
		\CheckmarkBold & \XSolidBrush & & \textbf{90.7} & \underline{83.9} & \textbf{89.9} & \underline{92.6} & \textbf{90.5} & \underline{93.8} & & \underline{67.93} & \textbf{94.22} & \underline{9.19} & 24.66 & \underline{73.70} & \underline{10.83} \\
		\CheckmarkBold & \CheckmarkBold & & \textbf{90.7} & \textbf{84.1} & \underline{88.9}& \underline{92.6} & \underline{89.7} & \textbf{94.1} & & \underline{68.39} & \underline{93.88} & \textbf{8.65} & \underline{24.86} & 73.44 & 12.64 \\ \bottomrule
	\end{tabular}
	\caption{Ablation study on detection and segmentation models. The baseline losses are CIoU for detection and SLS for segmentation. "\CheckmarkBold" indicates that our method is used, while "\XSolidBrush" means the original method is used.}
	\label{combined_table}
\end{table*}
%----------------------------------------------------------------------------------------------
Table 3 presents the results of various mask-based loss functions, including SLS \cite{28}, IoU, and Dice \cite{34}, for IRST segmentation. Our SDM$(\delta=0.5)$ loss not only achieved the best overall performance but also maintained a strong balance across datasets.

From the ablation experiments in Tables 2 and 3, we observed that, in the detection model, smaller $\delta$ in SDB loss led to better performance on IRSTD-1K, while a larger $\delta$ improved performance on SIRST-UAVB. This is likely due to the nature of the labels: BBoxes have greater IoU fluctuations than masks, making them more sensitive to $\delta$ changes. A larger $\delta$ also widens the fluctuation range of the dynamic impact coefficient $\beta$, which is particularly beneficial for small targets, such as SIRST-UAVB, by reducing Sloss fluctuations and improving accuracy. Thus, in detection models, $\delta$ should be selected based on target size, while in segmentation models, $\delta = 0.5$ consistently provides the best balance.

\subsection{Ablation Experiments on Multiple Models}
As shown in Table 4, our proposed PConv and SDB loss consistently enhances performance across a range of detection and segmentation networks, including EFLNet \cite{eflnet}, YOLOv5n \cite{22}, YOLOv8n-p2, DNANet \cite{27}, ISNet \cite{26}, and MSHNet. 
The combination of PConv and SDB loss achieved the highest $mAP50$ scores across all detection models, demonstrating superior detection capability. Significant gains in precision and recall, particularly in EFLNet, further validates the effectiveness of our approach in overcoming the limitations of traditional convolutional layers and loss functions. Overall, PConv and SDB loss functions offer clear advantages in enhancing detection accuracy, stability, and generalization, establishing them as powerful tools for boosting detection network performance.

In segmentation models, the combination of PConv with SDM loss consistently led to significant improvements, especially in DNANet, with notable gains also observed in ISNet and MSHNet. While MSHNet performed better than the baseline with the combined approach, it did not surpass the performance of PConv with SDM loss alone, indicating that optimal configurations may need to be tailored to specific architectures. Overall, our method proves robust and highly effective across different models.

We have further analyzed qualitative results for PConv and SD loss. As shown in Fig. 6 and Fig. 7, PConv reduces missed detections, while SD loss enhances weak signal detection. Together, they reduce false alarms and improve robustness.

\section{Conclusion}
In this paper, we proposed a plug-and-play PConv module, leveraging the Gaussian distribution characteristics of IRST to achieve an efficient, larger receptive field with minimal parameters. We also introduced a simple yet effective SD loss function to address IoU fluctuation issues with labels. Through extensive comparisons with existing convolutional modules and loss functions, our methods consistently outperformed state-of-the-art approaches, demonstrating superior accuracy and robustness. We validated the effectiveness and strong generalization capabilities of our approach across multiple models, showcasing its potential in advancing IRSTDS. Furthermore, we introduced the SIRST-UAVB dataset, a large-scale and challenging benchmark with detailed annotations.

\section{Acknowledgments}
We express our heartfelt thanks to the reviewers for their thoughtful feedback and insightful comments. We are grateful for the partial support from the Start-up Foundation for Ph.D. students at Southwest University of Science and Technology (No. 20zx7120).

\bibliography{aaai25}

\begin{thebibliography}{38}
\providecommand{\natexlab}[1]{#1}

\bibitem[{Chen et~al.(2021)Chen, Wang, Guo, Zhang, and Sun}]{drconv}
Chen, J.; Wang, X.; Guo, Z.; Zhang, X.; and Sun, J. 2021.
\newblock Dynamic region-aware convolution.
\newblock In \emph{Proceedings of the IEEE/CVF Conference on Computer Vision
  and Pattern Recognition}, 8064--8073.

\bibitem[{Chollet(2017)}]{sdconv}
Chollet, F. 2017.
\newblock Xception: Deep learning with depthwise separable convolutions.
\newblock In \emph{Proceedings of the IEEE Conference on Computer Vision and
  Pattern Recognition}, 1251--1258.

\bibitem[{Ciocarlan et~al.(2024)Ciocarlan, H{\'e}garat-Mascle, Lefebvre,
  Woiselle, and Barbanson}]{23}
Ciocarlan, A.; H{\'e}garat-Mascle, S.~L.; Lefebvre, S.; Woiselle, A.; and
  Barbanson, C. 2024.
\newblock A \textit{Contrario} Paradigm for YOLO-based Infrared Small Target
  Detection.
\newblock arXiv:2402.02288.

\bibitem[{Dai et~al.(2023)Dai, Li, Zhou, Qian, Chen, and Yang}]{oscar}
Dai, Y.; Li, X.; Zhou, F.; Qian, Y.; Chen, Y.; and Yang, J. 2023.
\newblock One-Stage Cascade Refinement Networks for Infrared Small Target
  Detection.
\newblock \emph{IEEE Transactions on Geoscience and Remote Sensing}, 61: 1--17.

\bibitem[{Dai et~al.(2021{\natexlab{a}})Dai, Wu, Zhou, and Barnard}]{25}
Dai, Y.; Wu, Y.; Zhou, F.; and Barnard, K. 2021{\natexlab{a}}.
\newblock Asymmetric contextual modulation for infrared small target detection.
\newblock In \emph{2021 IEEE Winter Conference on Applications of Computer
  Vision (WACV)}, 949--958.

\bibitem[{Dai et~al.(2021{\natexlab{b}})Dai, Wu, Zhou, and Barnard}]{12}
Dai, Y.; Wu, Y.; Zhou, F.; and Barnard, K. 2021{\natexlab{b}}.
\newblock Attentional local contrast networks for infrared small target
  detection.
\newblock \emph{IEEE Transactions on Geoscience and Remote Sensing}, 59(11):
  9813--9824.

\bibitem[{Deshpande et~al.(1999)Deshpande, Er, Venkateswarlu, and Chan}]{9}
Deshpande, S.~D.; Er, M.~H.; Venkateswarlu, R.; and Chan, P. 1999.
\newblock Max-mean and max-median filters for detection of small targets.
\newblock In \emph{Signal and Data Processing of Small Targets 1999}, volume
  3809, 74--83. SPIE.

\bibitem[{Du and Hamdulla(2019)}]{13}
Du, P.; and Hamdulla, A. 2019.
\newblock Infrared small target detection using homogeneity-weighted local
  contrast measure.
\newblock \emph{IEEE Geoscience and Remote Sensing Letters}, 17(3): 514--518.

\bibitem[{Jocher et~al.(2022)Jocher, Chaurasia, Stoken, Borovec, Kwon, Michael,
  Fang, Wong, Yifu, Montes et~al.}]{22}
Jocher, G.; Chaurasia, A.; Stoken, A.; Borovec, J.; Kwon, Y.; Michael, K.;
  Fang, J.; Wong, C.; Yifu, Z.; Montes, D.; et~al. 2022.
\newblock ultralytics/yolov5: v6. 2-yolov5 classification models, apple m1,
  reproducibility, clearml and deci. ai integrations.
\newblock \emph{Zenodo}.

\bibitem[{Karim and Andersson(2013)}]{2}
Karim, A.; and Andersson, J.~Y. 2013.
\newblock Infrared detectors: Advances, challenges and new technologies.
\newblock In \emph{IOP Conference Series: Materials Science and Engineering},
  volume~51, 012001. IOP Publishing.

\bibitem[{Kou et~al.(2022)Kou, Wang, Fu, Yu, and Zhang}]{6}
Kou, R.; Wang, C.; Fu, Q.; Yu, Y.; and Zhang, D. 2022.
\newblock Infrared small target detection based on the improved density peak
  global search and human visual local contrast mechanism.
\newblock \emph{IEEE Journal of Selected Topics in Applied Earth Observations
  and Remote Sensing}, 15: 6144--6157.

\bibitem[{Kou et~al.(2023)Kou, Wang, Peng, Zhao, Chen, Han, Huang, Yu, and
  Fu}]{7}
Kou, R.; Wang, C.; Peng, Z.; Zhao, Z.; Chen, Y.; Han, J.; Huang, F.; Yu, Y.;
  and Fu, Q. 2023.
\newblock Infrared small target segmentation networks: A survey.
\newblock \emph{Pattern Recognition}, 143: 109788.

\bibitem[{Li et~al.(2022)Li, Xiao, Wang, Wang, Lin, Li, An, and Guo}]{27}
Li, B.; Xiao, C.; Wang, L.; Wang, Y.; Lin, Z.; Li, M.; An, W.; and Guo, Y.
  2022.
\newblock Dense nested attention network for infrared small target detection.
\newblock \emph{IEEE Transactions on Image Processing}, 32: 1745--1758.

\bibitem[{Li and Shen(2023)}]{21}
Li, R.; and Shen, Y. 2023.
\newblock YOLOSR-IST: A deep learning method for small target detection in
  infrared remote sensing images based on super-resolution and YOLO.
\newblock \emph{Signal Processing}, 208: 108962.

\bibitem[{Li et~al.(2023)Li, Hou, Zheng, Cheng, Yang, and Li}]{lskconv}
Li, Y.; Hou, Q.; Zheng, Z.; Cheng, M.-M.; Yang, J.; and Li, X. 2023.
\newblock Large selective kernel network for remote sensing object detection.
\newblock In \emph{Proceedings of the IEEE/CVF International Conference on
  Computer Vision}, 16794--16805.

\bibitem[{Li et~al.(2016)Li, Zhang, Yu, Tan, Tian, and Ma}]{20}
Li, Y.; Zhang, Y.; Yu, J.-G.; Tan, Y.; Tian, J.; and Ma, J. 2016.
\newblock A novel spatio-temporal saliency approach for robust dim moving
  target detection from airborne infrared image sequences.
\newblock \emph{Information Sciences}, 369: 548--563.

\bibitem[{Liu et~al.(2024)Liu, Liu, Zheng, Wang, and Fu}]{28}
Liu, Q.; Liu, R.; Zheng, B.; Wang, H.; and Fu, Y. 2024.
\newblock Infrared Small Target Detection with Scale and Location Sensitivity.
\newblock arXiv:2403.19366.

\bibitem[{Liu et~al.(2023{\natexlab{a}})Liu, Yin, Yang, Wang, and An}]{5}
Liu, T.; Yin, Q.; Yang, J.; Wang, Y.; and An, W. 2023{\natexlab{a}}.
\newblock Combining deep denoiser and low-rank priors for infrared small target
  detection.
\newblock \emph{Pattern Recognition}, 135: 109184.

\bibitem[{Liu et~al.(2023{\natexlab{b}})Liu, Liu, Hao, Tang, Zhang, and
  Lei}]{19}
Liu, Y.; Liu, X.; Hao, X.; Tang, W.; Zhang, S.; and Lei, T. 2023{\natexlab{b}}.
\newblock Single-Frame Infrared Small Target Detection by High Local Variance,
  Low-Rank and Sparse Decomposition.
\newblock \emph{IEEE Transactions on Geoscience and Remote Sensing}.

\bibitem[{Luo et~al.(2016)Luo, Li, Urtasun, and Zemel}]{37}
Luo, W.; Li, Y.; Urtasun, R.; and Zemel, R. 2016.
\newblock Understanding the effective receptive field in deep convolutional
  neural networks.
\newblock \emph{Advances in Neural Information Processing Systems}, 29.

\bibitem[{Ma et~al.(2022)Ma, Yang, Wang, Sun, Ren, and Ahmad}]{3}
Ma, T.; Yang, Z.; Wang, J.; Sun, S.; Ren, X.; and Ahmad, U. 2022.
\newblock Infrared small target detection network with generate label and
  feature mapping.
\newblock \emph{IEEE Geoscience and Remote Sensing Letters}, 19: 1--5.

\bibitem[{Rezatofighi et~al.(2019)Rezatofighi, Tsoi, Gwak, Sadeghian, Reid, and
  Savarese}]{33}
Rezatofighi, H.; Tsoi, N.; Gwak, J.; Sadeghian, A.; Reid, I.; and Savarese, S.
  2019.
\newblock Generalized intersection over union: A metric and a loss for bounding
  box regression.
\newblock In \emph{Proceedings of the IEEE/CVF Conference on Computer Vision
  and Pattern Recognition}, 658--666.

\bibitem[{Rivest and Fortin(1996)}]{8}
Rivest, J.-F.; and Fortin, R. 1996.
\newblock Detection of dim targets in digital infrared imagery by morphological
  image processing.
\newblock \emph{Optical Engineering}, 35(7): 1886--1893.

\bibitem[{Sudre et~al.(2017)Sudre, Li, Vercauteren, Ourselin, and
  Jorge~Cardoso}]{34}
Sudre, C.~H.; Li, W.; Vercauteren, T.; Ourselin, S.; and Jorge~Cardoso, M.
  2017.
\newblock Generalised dice overlap as a deep learning loss function for highly
  unbalanced segmentations.
\newblock In \emph{Deep Learning in Medical Image Analysis and Multimodal
  Learning for Clinical Decision Support: Third International Workshop, DLMIA
  2017, and 7th International Workshop, ML-CDS 2017, Held in Conjunction with
  MICCAI 2017, Qu{\'e}bec City, QC, Canada, September 14, Proceedings 3},
  240--248. Springer.

\bibitem[{Sun(2024)}]{38}
Sun, H. 2024.
\newblock Ultra-High Resolution Segmentation via Boundary-Enhanced
  Patch-Merging Transformer.
\newblock arXiv:2412.10181.

\bibitem[{Sun, Yang, and An(2020)}]{4}
Sun, Y.; Yang, J.; and An, W. 2020.
\newblock Infrared dim and small target detection via multiple subspace
  learning and spatial-temporal patch-tensor model.
\newblock \emph{IEEE Transactions on Geoscience and Remote Sensing}, 59(5):
  3737--3752.

\bibitem[{Tan and Le(2019)}]{mixconv}
Tan, M.; and Le, Q.~V. 2019.
\newblock Mixconv: Mixed depthwise convolutional kernels.
\newblock arXiv:1907.09595.

\bibitem[{Wang et~al.(2021)Wang, Xu, Yang, and Yu}]{nwd}
Wang, J.; Xu, C.; Yang, W.; and Yu, L. 2021.
\newblock A normalized Gaussian Wasserstein distance for tiny object detection.
\newblock arXiv:2110.13389.

\bibitem[{Yang et~al.(2024)Yang, Zhang, Zhang, Luo, Zhou, and Pi}]{eflnet}
Yang, B.; Zhang, X.; Zhang, J.; Luo, J.; Zhou, M.; and Pi, Y. 2024.
\newblock EFLNet: Enhancing Feature Learning Network for Infrared Small Target
  Detection.
\newblock \emph{IEEE Transactions on Geoscience and Remote Sensing}, 62: 1--11.

\bibitem[{Ying et~al.(2024)Ying, Xiao, Li, He, Li, Li, Wang, Hu, Xu, Lin
  et~al.}]{safit}
Ying, X.; Xiao, C.; Li, R.; He, X.; Li, B.; Li, Z.; Wang, Y.; Hu, M.; Xu, Q.;
  Lin, Z.; et~al. 2024.
\newblock Visible-Thermal Tiny Object Detection: A Benchmark Dataset and
  Baselines.
\newblock arXiv:2406.14482.

\bibitem[{Yu and Koltun(2015)}]{dconv}
Yu, F.; and Koltun, V. 2015.
\newblock Multi-scale context aggregation by dilated convolutions.
\newblock arXiv:1511.07122.

\bibitem[{Zhang et~al.(2022)Zhang, Zhang, Yang, Bai, Zhang, and Guo}]{26}
Zhang, M.; Zhang, R.; Yang, Y.; Bai, H.; Zhang, J.; and Guo, J. 2022.
\newblock ISNet: Shape matters for infrared small target detection.
\newblock In \emph{2022 IEEE/CVF Conference on Computer Vision and Pattern
  Recognition (CVPR)}, 867--876.

\bibitem[{Zhang et~al.(2017)Zhang, Qi, Xiao, and Wang}]{gconv}
Zhang, T.; Qi, G.-J.; Xiao, B.; and Wang, J. 2017.
\newblock Interleaved group convolutions.
\newblock In \emph{Proceedings of the IEEE International Conference on Computer
  Vision}, 4373--4382.

\bibitem[{Zhang, Cong, and Wang(2003)}]{36}
Zhang, W.; Cong, M.; and Wang, L. 2003.
\newblock Algorithms for optical weak small targets detection and tracking.
\newblock In \emph{International Conference on Neural Networks and Signal
  Processing, 2003. Proceedings of the 2003}, volume~1, 643--647. IEEE.

\bibitem[{Zhang et~al.(2023)Zhang, Song, Song, Yang, Ye, Zhou, and
  Zhang}]{akconv}
Zhang, X.; Song, Y.; Song, T.; Yang, D.; Ye, Y.; Zhou, J.; and Zhang, L. 2023.
\newblock AKConv: Convolutional kernel with arbitrary sampled shapes and
  arbitrary number of parameters.
\newblock arXiv:2311.11587.

\bibitem[{Zhao et~al.(2022)Zhao, Li, Li, Hu, Ma, and Tao}]{1}
Zhao, M.; Li, W.; Li, L.; Hu, J.; Ma, P.; and Tao, R. 2022.
\newblock Single-frame infrared small-target detection: A survey.
\newblock \emph{IEEE Geoscience and Remote Sensing Magazine}, 10(2): 87--119.

\bibitem[{Zheng et~al.(2020)Zheng, Wang, Liu, Li, Ye, and Ren}]{31}
Zheng, Z.; Wang, P.; Liu, W.; Li, J.; Ye, R.; and Ren, D. 2020.
\newblock Distance-IoU loss: Faster and better learning for bounding box
  regression.
\newblock In \emph{Proceedings of the AAAI Conference on Artificial
  Intelligence}, volume~34, 12993--13000.

\bibitem[{Zhou et~al.(2023)Zhou, Li, Zhang, Lu, and Hu}]{18}
Zhou, X.; Li, P.; Zhang, Y.; Lu, X.; and Hu, Y. 2023.
\newblock Deep Low-Rank and Sparse Patch-Image Network for Infrared Dim and
  Small Target Detection.
\newblock \emph{IEEE Transactions on Geoscience and Remote Sensing}, 61: 1--14.

\end{thebibliography}

\appendix
\section{Appendix}

\subsection{Supplementary Experiment}
\begin{table}[h]
	\centering
	\begin{tabular}{c@{\hskip 0.1in}|@{\hskip 0.1in}c@{\hskip 0.1in}|@{\hskip 0.1in}c@{\hskip 0.1in}|@{\hskip 0.1in}c@{\hskip 0.1in}c@{\hskip 0.1in}c}
		\toprule
		Scale   & PConv & SDB & $P$     & $R$     & $mAP50$   \\ \midrule
		\multirow{8}{*}{(0,16]} & -     & -        & 0.906 & 0.871 & 0.913 \\
		& (3,3) & -        & 0.93  & 0.872 & 0.913 \\
		& (4,3) & -        & 0.936 & 0.884 & 0.923 \\
		& (4,4) & -        & 0.938 & 0.87  & 0.921 \\
		& -     & $\delta=0.3$      & 0.93  & 0.882 & 0.932 \\
		& -     & $\delta=0.5$      & 0.941 & 0.886 & 0.933 \\
		& -     & $\delta=0.7$      & 0.926 & 0.893 & 0.927 \\
		& (4,3) & $\delta=0.5$      & 0.940 & 0.882 & 0.924 \\  \midrule
		\multirow{8}{*}{(16,49]} & -     & -        & 0.931 & 0.898 & 0.953 \\ 
		& (3,3) & -        & 0.922 & 0.92  & 0.955 \\
		& (4,3) & -        & 0.922 & 0.929 & 0.956 \\
		& (4,4) & -        & 0.934 & 0.92  & 0.952 \\
		& -     & $\delta=0.3$      & 0.936 & 0.932 & 0.96  \\
		& -     & $\delta=0.5$      & 0.922 & 0.906 & 0.947 \\
		& -     & $\delta=0.7$      & 0.92  & 0.921 & 0.953 \\
		& (4,3) & $\delta=0.5$      & 0.95  & 0.886 & 0.952 \\  \midrule
		\multirow{8}{*}{(49,$\infty$)} & -     & -        & 0.983 & 1     & 0.995 \\ 
		& (3,3) & -        & 0.989 & 1     & 0.995 \\
		& (4,3) & -        & 0.986 & 1     & 0.995 \\
		& (4,4) & -        & 0.934 & 0.92  & 0.952 \\
		& -     & $\delta=0.3$      & 0.989 & 1     & 0.995 \\
		& -     & $\delta=0.5$      & 0.984 & 0.993 & 0.995 \\
		& -     & $\delta=0.7$      & 0.989 & 1     & 0.995 \\
		& (4,3) & $\delta=0.5$      & 0.99  & 1 & 0.995 \\ \bottomrule
	\end{tabular}
	\caption{The ablation experiment of PConv and SDB loss for different scale target detection performance.}
\end{table}
We divided the SIRST-UAVB dataset according to target scale and conducted comprehensive ablation experiments on PConv and SDB loss under the framework of YOLOv8n-p2. From Table 5, it can be observed that PConv and SDB loss enhance the detection capability across various target scales, particularly for small targets with box areas less than 16, effectively improving multiple metrics. Setting the convolution kernels of two PConv layers to 4 and 3, respectively, provides stable and significant performance improvements. Interestingly, the larger the combination of PConv kernel sizes, the better the detection accuracy performance for small targets. Conversely, the detection performance for large targets decreases. Setting the SDB loss parameter to $(\delta=0.5)$ enhances the detection performance for small targets. Additionally, combining PConv with SDB loss notably improves accuracy.

\subsection{Limitation} The main limitation is that the mask-based label SDM loss may not support setting too large batch size, because the mask label-based SLS loss is obtained by averaging the loss of batch size. The target size we calculated is also the average value based on batch size, but that in batch size may vary greatly, so the SDM loss does not really assign Sloss and Lloss influence coefficients according to the target size. However, too large batch size will aggravate this negative effect. In the future work, we will try to solve this problem by assigning the correct influence coefficient according to the size of each target.

\end{document}